\documentclass[12pt,a4paper]{article}

\usepackage[utf8]{inputenc} 
\usepackage{graphicx} 
\usepackage[colorinlistoftodos]{todonotes}

\usepackage{booktabs} 
\usepackage{array} 
\usepackage{paralist} 
\usepackage{verbatim} 
\usepackage{subfig} 
\usepackage{caption}
\usepackage{subcaption}

\usepackage[toc,page]{appendix}
\usepackage{pdflscape}
\usepackage{enumitem}
\usepackage{amssymb}
\usepackage{calc}

\usepackage{geometry}

\usepackage{optidef}
\usepackage{multirow}
\usepackage{xcolor}
\usepackage{pdflscape}
\usepackage{algorithm}
\usepackage{algpseudocode}
\usepackage{listings}
\usepackage{wrapfig}
\usepackage{ragged2e}
\usepackage{calligra}
\usepackage[T1]{fontenc}
\usepackage[bb=boondox]{mathalfa}
\usepackage[hidelinks]{hyperref}
\usepackage{authblk}

\author{Neil Urquhart}
\author{Amir(Navid) Rahimi}
\author{Efstathios-Al. Tingas}
\affil{School of Computing, Engineering and the Built Environment, Edinburgh Napier University, Edinburgh, EH10 5DT, UK
n.urquhart,a.rahimi,e.tingas @napier.ac.uk}
 
\title{Optimisation of Aircraft Maintenance Schedules}

\begin{document}
\maketitle
\begin{abstract}
We present an aircraft maintenance scheduling problem, which requires suitably qualified staff to be assigned to maintenance tasks on each aircraft. The tasks on each aircraft must be completed within a given turn around window so that the aircraft may resume revenue earning service. This paper presents an initial study based on the  application of an Evolutionary Algorithm to the problem. Evolutionary Algorithms evolve a solution to a problem by evaluating many possible solutions, focusing the search on those solutions that are of a higher quality, as defined by a fitness function. In this paper, we benchmark the algorithm on 60 generated problem instances to demonstrate the underlying representation and associated genetic operators. 

\end{abstract}

\section{Introduction}


Aircraft maintenance scheduling is a critical component of aviation operations, as it directly impacts flight safety, operational efficiency, and cost management. Airlines must carefully balance routine maintenance checks, unplanned repairs, and fleet utilisation to ensure aircraft remain operational while minimising downtime. The interdependence of these factors makes maintenance scheduling an inherently complex problem, requiring meticulous coordination among multiple stakeholders, including maintenance crews, airport ground services, airline operations teams, and regulatory bodies.

Effective maintenance scheduling extends beyond simple time-based planning and involves integrating maintenance tasks with aircraft routing, crew availability, and airport constraints. Aircraft must be routed through designated maintenance hubs at strategically planned intervals to minimise disruptions to scheduled flights while ensuring compliance with stringent regulatory requirements. This interplay between scheduling, maintenance, and operational logistics requires sophisticated optimisation techniques that can accommodate real-time operational constraints, flight delays, and unexpected maintenance needs. Well-optimised maintenance scheduling significantly reduces operational costs by preventing last-minute aircraft groundings, mitigating disruptions, and optimising component replacements, thereby extending fleet longevity and maintaining high levels of service reliability.

Given the scale and complexity of aircraft maintenance scheduling, mathematical optimisation techniques have played a crucial role in developing efficient scheduling frameworks. Integer Linear Programming (ILP) and column generation heuristics \cite{CACCHIANI2013391} have been widely employed to decompose large-scale maintenance scheduling problems into smaller, more manageable subproblems, allowing for more efficient computation and decision-making. In \cite{CUI2019106045}, heuristic-based approaches further enhance computational efficiency by focusing on solution feasibility while reducing the computational burden associated with traditional mathematical models. Similarly, \cite{diaz2014aircraft} explores the integration of multiple heuristic strategies to refine maintenance planning, demonstrating how hybrid approaches can be adapted to meet industry-specific constraints and improve scheduling flexibility.

As the aviation sector continues to evolve, there is an increasing emphasis on incorporating advanced optimisation techniques, real-time data analytics, and predictive maintenance models to enhance operational resilience and efficiency. The integration of artificial intelligence (AI) and machine learning (ML) into maintenance scheduling represents a significant breakthrough in the industry. AI-driven predictive maintenance models leverage vast amounts of operational data, sensor readings, and historical maintenance records to anticipate potential failures before they occur. This predictive approach allows airlines to transition from reactive or preventive maintenance strategies to condition-based and proactive maintenance planning, optimising resource allocation and reducing unexpected downtime.

In this paper we investigate the use of an evolutionary algorithm to build maintenance schedules by allocating staff, and where possible, determining the order of work.  In the problem under consideration each aircraft requiring maintenance has a strict time window, known as a \textit{turnaround window} within which a series of \textit{work packages} must be scheduled. The order of the work packages may be changed as long as all of the work packages are completed within the turn around window. Each work package comprises one or more \textit{work orders}, each of which specifies a operation to be carried out, a duration and the number of technicians required. Each technician may be required to hold a specific qualification to undertake that operation. The ordering of work orders within a work package may not be changed.

This paper is organised as follows, in section 2 we present a review of relevant previous work, we present an evolutionary algorithm (known as \textit{AERO-EA}) in section 3. Results obtained through the application of AERO-EA to a set of problems produced using a problem generator are given in section 4.

\section{Previous Work}


Meta-heuristic algorithms, particularly evolutionary approaches, have gained widespread adoption across various industrial sectors due to their effectiveness in tackling complex optimisation problems. These include timetabling \cite{Atiq-2021,chen2020tabu}, staff scheduling \cite{Burke-2013,abdelghany_2021,si-2022}, vehicle routing and logistics \cite{kent-2014,Kondratenko-2021}, and job shop/factory scheduling \cite{Kittel-2021}. Such problems fall within the realm of combinatorial optimisation, where the search for optimal or near-optimal solutions is computationally intensive due to the vast number of possible configurations and constraints involved.

Meta-heuristic optimization techniques are particularly advantageous in scenarios where exact methods become impractical due to the high dimensionality of the search space. These algorithms provide approximate solutions by intelligently navigating complex landscapes using heuristics that balance exploration (searching new regions) and exploitation (refining promising solutions). Solution methodologies typically involve generating candidate solutions and assessing their quality through a predefined fitness function. Some methods iteratively refine a single solution through local search techniques, whereas population-based approaches, particularly those employing evolutionary algorithms, maintain a diverse set of solutions that evolve over successive generations.

Evolutionary algorithms, inspired by biological evolution, operate through mechanisms such as selection, crossover, and mutation. These techniques ensure that high-quality solutions are preserved while introducing diversity into the population, preventing premature convergence to suboptimal results. Classic evolutionary strategies, such as Genetic Algorithms (GA) and Differential Evolution (DE), have demonstrated significant success in optimising highly nonlinear and constrained problems across industries.

Traditional evolutionary algorithms primarily focus on converging toward a single globally optimal solution. However, recent advancements in the field have introduced paradigms such as Quality Diversity (QD) \cite{pugh2016quality} and Illumination Algorithms \cite{mouret2015illuminating}. These approaches prioritize not just optimization but also the discovery of a diverse set of high-performing solutions across different regions of the solution space. By promoting diversity, these algorithms enable decision-makers to explore a spectrum of viable alternatives, selecting solutions that best align with specific application requirements and constraints. This paradigm shift has expanded the applicability of evolutionary methods to complex, multi-objective, and dynamic optimization problems, where flexibility and adaptability are critical.

The integration of meta-heuristic algorithms with modern machine learning techniques further enhances their effectiveness. Hybrid models that combine evolutionary algorithms with deep learning or reinforcement learning allow for adaptive optimization strategies that learn from past solutions and improve performance over time. Additionally, parallel computing and cloud-based implementations have enabled large-scale meta-heuristic optimizations to be executed more efficiently, making them suitable for real-time applications in scheduling, logistics, and industrial automation.

As research in meta-heuristics continues to evolve, future directions may include further refinements in adaptive heuristics, self-tuning evolutionary strategies, and their integration with advanced AI-driven optimization frameworks. By leveraging the strengths of evolutionary computation alongside emerging computational intelligence techniques, these methods will continue to play a crucial role in solving increasingly complex industrial and scientific problems.


The Workforce Routing and Scheduling Problem (WSRP) \cite{castillo-salazar-2012} involves coordinating a set of mobile workers who must travel between multiple locations to complete assigned tasks. WSRP instances arise in various industries, including health and social home-care scheduling \cite{hiermann-2013,urquhart2018optimisation} and technician scheduling \cite{Hart-2014}. A bi/multi-objective formulation of the WSRP is presented in \cite{Braekers-2016}, where cost and patient convenience serve as the twin objectives. In \cite{hiermann-2013}, a healthcare WSRP is modelled as a single-objective problem that minimises working and travel time while penalising violations of hard and soft constraints. This formulation also incorporates a binary transport mode choice (car or public transport). In \cite{Braekers-2016}, the trade-off between cost and client inconvenience is further explored in a home-care scheduling context. Approaches to solving the WSRP include meta-heuristics \cite{bertels-2006} and hyper-heuristics \cite{Hart-2014}. For a comprehensive survey of applied methods, the reader is directed to \cite{castillo-salazar-2012}.

The scheduling of healthcare professionals has been extensively studied, driven by the need to optimise healthcare provision and ensure the efficient utilisation of highly trained staff. Healthcare scheduling problems typically involve shift constraints and staff qualification requirements, as only certain personnel are qualified to perform specific tasks. The authors of \cite{si-2022} employ an Evolutionary Algorithm (EA) to schedule staff in a distribution centre. The EA assigns groups of tasks to staff members, ensuring that all work is covered while maintaining feasible allocations for individuals. This problem domain is characterised by a large number of inflexible, time-dependent tasks.

Significant research has been conducted on scheduling healthcare workers, a domain that shares key similarities with the current study, particularly in handling soft constraints and shift-based scheduling. For a recent survey of healthcare workforce scheduling, the reader is referred to \cite{ngoo_2022}. 

In \cite{Burke-2013}, a variable-depth search algorithm is proposed for the nurse rostering problem. This method constructs sequences of neighbourhood swaps, employing heuristics to form effective move chains. The study specifically addresses shift rostering and its staffing implications. Similarly, \cite{abdelghany_2021} investigates related problem instances using a neighbourhood search approach to generate an initial feasible solution. This initial solution is subsequently refined in a second phase, where further swaps are applied to enhance the overall schedule quality.

\section{Methodology}
\subsection{Problem Formulation}

We consider the problem of optimally assigning certified maintenance technicians to multiple aircraft undergoing service within fixed turnaround intervals. Each aircraft has a fixed turnaround window, defined by a scheduled landing and departure time, within which all required maintenance tasks must be completed. These tasks are described in terms of work packages (WPs), which serve as high-level groupings of more granular work orders (WOs). The assignment must satisfy a range of temporal, resource, and qualification constraints.

Each WP encompasses one or more WOs, which are atomic maintenance tasks requiring uninterrupted execution. While WPs are independent and may be executed in any order, all WOs within a given WP must be completed sequentially. That is, there is an intrinsic temporal dependency at the work order level: the next WO in the same WP cannot begin until the preceding one finishes. This leads to a hierarchical temporal structure, where inter-package scheduling is flexible, but intra-package execution is constrained by WO sequences.

The task pool comprises 24 unique WPs, each described by: (i) its estimated duration in minutes, ranging from 45 minutes to 150 minutes; (ii) the total number of man-hours required to complete the task; (iii) the number of technicians or engineers required for concurrent execution; and (iv) the specific certification(s) required. Certification requirements are heterogeneous: while some WPs can be performed by either B1 Technician or B2 Technician, others demand higher-grade certifications such as B1 Engineer or B2 Engineer, and several require dual certifications (e.g., B1 Engineer, B2 Engineer). This imposes non-trivial constraints on technician eligibility.

The technician pool includes 36 personnel, each possessing one of the four recognised certification types: B1 Technician, B2 Technician, B1 Engineer, or B2 Engineer. The distribution of these certifications is non-uniform, creating an additional layer of constraint on resource availability. Each technician is available for a single shift of fixed duration (e.g., 8 hours) and cannot be allocated to overlapping tasks. A technician is considered assigned if they meet the certification requirements for a WP and are available continuously throughout the required task duration.

The key decision variables in the problem are binary indicators denoting whether a technician is assigned to a given WP on a given aircraft. The objective is to find a feasible and efficient assignment of technicians such that:
\begin{itemize}
\item All WPs for all aircraft are completed within the respective turnaround windows;
\item Technician assignments satisfy certification constraints and resource requirements (number of personnel);
\item WOs are executed in correct sequential order where applicable;
\item No technician is double-booked, and workload is distributed evenly to the extent possible.
\end{itemize}

The problem is inherently combinatorial and multi-constrained, involving mixed qualification logic, temporal sequencing, and personnel capacity limits. Due to the complexity of overlapping aircraft schedules, non-pre-emptive task execution, and multi-skill matching, this problem falls within the class of NP-hard scheduling problems and is thus well-suited for meta-heuristic optimisation techniques such as genetic algorithms.

\subsection{Data Generation}

The authors developed a problem generator that produces  problem instances that represent realistic aircraft maintenance schedules over a 24-hour operational window. Each problem instance includes maintenance data for 20 aircraft, for each aircraft we specify the aircraft serial number, landing date and time, departure date and time, computed turnaround duration, and a list of work packages (WPs) assigned to that aircraft. The CSV format is directly usable as input for the optimisation algorithm, enabling systematic testing of technician allocation under varying workload and timing conditions.

Each generated instance is guaranteed to satisfy three key constraints that capture operational and planning realities. First, coverage: all 24 defined WPs must appear at least once across the full fleet of 20 aircraft. This ensures that the optimisation problem includes the complete diversity of maintenance tasks and certification requirements. Second, workload balance: the cumulative man-hours required by all tasks assigned in the instance must lie within a target window typically between 70\% and 90\% of the total available capacity, which is computed as the number of technicians (36) multiplied by their shift duration (8 hours). This avoids trivial under loaded or overloaded scenarios, ensuring the optimisation operates under realistic pressure. Third, turnaround feasibility: the computed turnaround time for each aircraft is defined as a scaled multiple of the total task duration. This scaling factor serves a dual role. It determines how long the aircraft is grounded relative to the time required to perform the assigned maintenance, effectively setting the operational time window available for technician assignment. For example, a factor of 1.2 corresponds to a 20\% buffer beyond the raw task time. This approach adds realism by capturing variability in ground time due to scheduling flexibility, maintenance uncertainty, or other operational considerations. The total turnaround duration is subject to a strict upper limit of 28 hours to reflect practical constraints on aircraft availability.

The problem generation commences by loading the work package definitions from a CSV file. The generation process then proceeds instance-by-instance. For each instance, the algorithm randomly selects a base day and attempts to construct a full 20-aircraft schedule that satisfies all constraints. Each aircraft is assigned a random landing time within the 24-hour day. To ensure WP coverage, the generator first assigns at least one WP that has not yet been used in the current instance. It then stochastically adds further WPs to each aircraft from the full WP list, allowing repetition. This process is controlled by a soft condition: once the cumulative task duration exceeds 20 minutes and a probabilistic stop is triggered, or after 50 retries, WP selection ends.

After each aircraft’s list of WPs is finalised, the total task time is scaled by the turnaround factor (e.g. 1.2) to compute the turnaround duration. The aircraft’s departure time is then calculated as its landing time plus this turnaround. If any aircraft exceeds the maximum allowed turnaround time, the attempt is invalidated and restarted. Once all 20 aircraft are populated, we check whether all WPs are covered and whether the total man-hour usage falls within the permitted range. If these checks fail, the attempt is discarded and a new one begins. This process is repeated up to 1000 times per instance to ensure success. Upon satisfaction of all constraints, the instance is written to a CSV file. 
The code supports full reproducibility via an optional random seed, and its parameters—such as the number of aircraft, technicians, and the turnaround scaling factor—can be adjusted to explore different load scenarios. The result is a robust framework for generating high-fidelity, constraint-compliant data that reflects real-world aircraft maintenance planning challenges.

Table~\ref{tab:probs} summarises the two batches of problem instances generated for use in this paper. In total, 60 instances were created, each comprising 20 aircraft. The two batches differ in the scaling factor applied to the cumulative work package (WP) duration when computing aircraft turnaround time. This factor directly controls the temporal flexibility available to schedule maintenance tasks. In Batch (a), 40 instances were generated using a high scaling factor such that, on average, the total duration of WPs accounted for only 35\% of the allocated turnaround window. This setting simulates operational conditions with ample slack, allowing the optimisation to explore a wider range of technician allocations. In contrast, Batch (b) comprises 20 instances in which the WP duration consumes, on average, 83\% of the turnaround time. This creates a more constrained environment, emulating tighter ground time conditions often encountered in high-throughput or delay-sensitive scenarios. The table reports key statistics across both batches, including the average total turnaround time per instance, average cumulative WP duration, and the average number of WPs assigned per aircraft. These metrics highlight the effect of turnaround compression on workload density and underscore the differing levels of scheduling difficulty introduced in each batch.
\begin{table}[]
\tiny
\centering
\begin{tabular}{ccccccc}
\toprule
Batch & Instances & \begin{tabular}[c]{@{}c@{}}Aircraft \\ per instance\end{tabular} & \begin{tabular}[c]{@{}c@{}}Avg. Total Turnaround\\  time (mins)\end{tabular} & \begin{tabular}[c]{@{}c@{}}Avg. WP\\ Total Duration (mins)\end{tabular} & \begin{tabular}[c]{@{}c@{}}\% of Turnaround \\ need for WPs\end{tabular} & \begin{tabular}[c]{@{}c@{}}Avg WPs per\\ Aircraft\end{tabular} \\
\midrule
a     & 40        & 20                                                               & 14487.3                                                                      & 5142.75                                                                 & 35\%                                                                     & 2.7                                                            \\
b     & 20        & 20                                                               & 6021.6                                                                       & 5018                                                                    & 83\%                                                                      & 2.3     \\   
\bottomrule
\end{tabular}
\caption{In total 60 problem instances were generated, in the first batch of 40 problem instances (a) the length of time needed for maintenance was on average only 35\% of the turnaround window. In the second batch (b) this was increased to 83\%}
\label{tab:probs}
\end{table}

\subsection{The AERO-EA Algorithm}
\subsubsection{General Description}

AERO-EA is a steady-state EA, which produces one child per generation. Each child is created by either cloning a single parent or the recombination of two parents, the operator being selected at random.  Parent selection is by a tournament of size $ks$ (see table \ref{tab:params}). 

The mutation operator is then applied to each child in order to introduce some random variance into the child chromosome. The child is then evaluated and allocated a fitness value. A replacement tournament operator (of size $kr$) is applied to the population. In this case the individual with the worst fitness is selected. If the fitness of the selected individual is worse that that of the child then the child replaces that individual.  This tournament based replacement strategy protects the best individual in the population from replacement. 

The algorithm runs for a set number of evaluations as specified by the evaluation budget (see table \ref{tab:params}).

\subsubsection{Representation and Evaluation}

The genotype structure must be capable of allowing work packages to be re-ordered whilst preserving the immutability of the work orders within each work package.  Each chromosome is a permutation of genes, where a gene represents one work package associated with an aircraft. For example, table \ref{tab:chromo} shows a solution based on a problem comprising 3 aircraft $a_0$... $a_3$ with 4 possible work packages $wp_0$ ... $wp_3$ allocated to them.

\begin{table}[]
\tiny
\centering
\begin{tabular}{|l|}
\hline
a0.wp1 \\ \hline
a1.wp3 \\ \hline
a2.wp0 \\ \hline
a0.wp0 \\ \hline
a1.wp4 \\ \hline
a2.wp2 \\ \hline
\end{tabular}
    \caption{A sample chromosome, comprising a permutation of genes, each gene representing one instance of an aircraft and associated work package.}
    \label{tab:chromo}
\end{table}
The permutation can be reordered, at the decoding stage the work packages are added to the solution in the given order.  Each gene (work package) contains the set of work orders associated with that work package. The first gene in the above chromosome (aircraft 1, work package 1) could be expanded as follows to show the work orders:\\

\begin{table}[]
\tiny
\centering
\begin{tabular}{c|c|c|}
\cline{2-3}
\multirow{6}{*}{a0.wp1 $\rightarrow$}& a0.wp1.wo1.techA & \textit{s0} \\ \cline{2-3} 
                        & a0.wp1.wo1.techA & \textit{s1} \\ \cline{2-3} 
                        & a0.wp1.wo1.techC & \textit{s2} \\ \cline{2-3} 
                        & a0.wp1.wo2.techB & \textit{s3} \\ \cline{2-3} 
                        & a0.wp1.wo3.techA & \textit{s0} \\ \cline{2-3} 
                        & a0.wp1.wo3.techB & \textit{s3} \\ \cline{2-3} 
\end{tabular}
    \caption{The a0.wp1 gene (see table \ref{tab:chromo}) showing three work orders and the 6 staff allocations needed to fulfil these work orders.}
    \label{tab:gene}
\end{table}

In example in table \ref{tab:chromo} $wp1$ comprises 3 work orders, which need 3,1 and 2 staff respectively to fulfil them.  Table \ref{tab:gene} shows the 6 entries within the gene, one for each member of staff required.
The EA operators are able to alter the order of the work packages and the staff allocated to the work orders.

In order to evaluate an individual the chromosome is decoded into a schedule. Work packages are added into the solution in the order they appear in the chromosome. If a work package is the first to be added to be added for an aircraft, the default commencement time will be start of the aircraft turnaround window, otherwise it will be based on the finish time of the previous work package.  Each work order within that package, has their default time allocated based on the finishing time of the previous work order. Each entry in the gene is considered and it is attempted to allocate the suggested member of staff to the work order, if they are available at that time (a member of staff may have already been allocated work that conflicts). If the member of staff cannot be allocated due to a conflict then that item of work is left uncovered and penalised (see below). If a member of staff is available at a later time, and the work order (and all subsequent work orders in the package) can be moved forward to accommodate then the start times will be updated. Constraints in the initialisation, mutation and crossover operators ensure that staff are only allocated to work they are qualified for.

At the end of the decoding process the fitness of the solution is calculated based on a penalty function as follows $$f=(w*wp)+(l*lp)$$
where:
\begin{itemize}
\item[] $f$ is the fitness, the higher the value of $f$ the worse the solution
\item[] $w$ is the number of missing members of staff (i.e. the member of staff was not allocated due to a time conflict)
\item[] $wp$ the penalty value for an item of work being uncovered
\item[] $l$ is the number of aircraft where the maintenance work concludes after end of their turn around window
\item[] $lp$ is the penalty value associated with breaking a turnaround window
\end{itemize}

Values for $wp$ and $lp$ are given in table \ref{tab:params}.

\subsubsection{Operators}

Each member of the population is initialised by randomly ordering the genome (which determines the order in which work packages are added to the schedule), within each gene, members of staff are allocated randomly, subject to the constraint that only members of staff with appropriate qualification are considered. 

The crossover operator creates a child based on the  crossover  of two parents (see table \ref{tab:xo}). Genes are copied to the  child by considering the first gene in \textit{Pa} ($s0.wp1$) then the first gene in \textit{Pb} ($a2.wp0$), then the next in \textit{Pa} ($1.wp3$) if the gene has not already been added to the child then it is added. This crossover operator preserves the priority of genes (e.g. $a0.wp1$ and $a2.wp0$) remain at the start of the child and so are still a high priority for adding to the schedule.  It should be remembered that the allocation of staff to work packages (see table \ref{tab:gene}) is also preserved and so the child inherits staff allocations from both parents.

\begin{table}[]
\centering
\tiny
\begin{tabular}{clcll}
\textit{Pa}                  &                       & \textit{Pb}                 &                       & \textit{Child}              \\ \cline{1-1} \cline{3-3} \cline{5-5} 
\multicolumn{1}{|c|}{a0.wp1} & \multicolumn{1}{l|}{} & \multicolumn{1}{c|}{a2.wp0} & \multicolumn{1}{l|}{} & \multicolumn{1}{l|}{a0.wp1} \\ \cline{1-1} \cline{3-3} \cline{5-5} 
\multicolumn{1}{|c|}{a1.wp3} & \multicolumn{1}{l|}{} & \multicolumn{1}{c|}{a0.wp0} & \multicolumn{1}{l|}{} & \multicolumn{1}{l|}{a2.wp0} \\ \cline{1-1} \cline{3-3} \cline{5-5} 
\multicolumn{1}{|c|}{a2.wp0} & \multicolumn{1}{l|}{} & \multicolumn{1}{c|}{a0.wp1} & \multicolumn{1}{l|}{} & \multicolumn{1}{l|}{a1.wp3} \\ \cline{1-1} \cline{3-3} \cline{5-5} 
\multicolumn{1}{|c|}{a0.wp0} & \multicolumn{1}{l|}{} & \multicolumn{1}{c|}{a1.wp4} & \multicolumn{1}{l|}{} & \multicolumn{1}{l|}{a0.wp0} \\ \cline{1-1} \cline{3-3} \cline{5-5} 
\multicolumn{1}{|c|}{a1.wp4} & \multicolumn{1}{l|}{} & \multicolumn{1}{c|}{a1.wp3} & \multicolumn{1}{l|}{} & \multicolumn{1}{l|}{a1.wp4} \\ \cline{1-1} \cline{3-3} \cline{5-5} 
\multicolumn{1}{|c|}{a2.wp2} & \multicolumn{1}{l|}{} & \multicolumn{1}{c|}{a2.wp2} & \multicolumn{1}{l|}{} & \multicolumn{1}{l|}{a2.wp2} \\ \cline{1-1} \cline{3-3} \cline{5-5} \\
\end{tabular}
\caption{The \textit{Child} chromosome is created based on features from both parents \textit{Pa} and \textit{Pb} as described in the text. Note that when each gene is copied to the child from either parent the gene also includes the staff allocations associated with that gene.}
\label{tab:xo}
\end{table}

Two mutation operators are utilised with an equal probability in \textit{AERO-EA} as follows:
\begin{itemize}
    \item An entry is randomly selected from within a randomly selected gene and randomly re-assigned to another qualified staff member.
    \item A gene is selected at random and moved to a randomly selected position within the chromosome.
\end{itemize}

\subsubsection{Parameters and Experimental Setup}

\textit{AERO-EA} was coded in Python 3 within a Jupyter Notebook.  To account for the stochastic nature of the EA, all runs were repeated 10 times.  

The parameters used can be seen in table \ref{tab:params}. For the purposes of experimentation the evaluation budget was set to a value that allowed for sufficient exploration of the solution space to find a solution with a suitably low fitness.  The values of the remaining parameters were  arrived at through empirical experimentation.

\begin{table}[]
\centering
\tiny
\begin{tabular}{c|c}
\toprule
Parameter    & Value  \\
\midrule
pop size     & 1500   \\
evals budget & 200000 \\
$ks$           & 2      \\
$kr$           & 2      \\
$lp$           & 10     \\
$wp$           & 1     \\
\bottomrule
\end{tabular}
\caption{The parameters used to generate the results given in section \ref{sec:res}.}
\label{tab:params}
\end{table}

\section{Results and Conclusions}

\subsection{Summary of Results}
\label{sec:res}
The AERO-EA algorithm was rigorously evaluated on a diverse set of 60 synthetically generated aircraft maintenance scheduling instances, systematically varying the temporal flexibility and workload density to emulate both relaxed and highly constrained operational scenarios (as outlined in Tables \ref{tab:res1} and \ref{tab:res2}). For each instance, the evolutionary search was repeated 10 times to assess stochastic performance, stability, and robustness.

\begin{table}[h]
    \centering
    \tiny
\begin{tabular}{l|ccccc}
\toprule
 Instance & Best & Avg. & Missing Staff &Late & \% Evals \\
\midrule
a1 & 0 & 1.0 & 1.0 & 0 & 70.6 \\
a2 & 0 & 1.5 & 1.5 & 0 & 62.3 \\
a3 & 0 & 0.7 & 0.7 & 0 & 70.1 \\
a4 & 0 & 0.6 & 0.6 & 0 & 70.2 \\
a5 & 0 & 1.0 & 1.0 & 0 & 61.1 \\
a6 & 0 & 0.9 & 0.9 & 0 & 71.0 \\
a7 & 0 & 0.3 & 0.3 & 0 & 51.7 \\
a8 & 0 & 0.3 & 0.3 & 0 & 85.3 \\
a9 & 0 & 0.6 & 0.6 & 0 & 62.4 \\
a10 & 0 & 0.1 & 0.1 & 0 & 35.8 \\
a11 & 0 & 0.4 & 0.4 & 0 & 64.7 \\
a12 & 0 & 0.0 & 0.0 & 0 & 59.5 \\
a13 & 0 & 2.1 & 2.1 & 0 & 82.5 \\
a14 & 0 & 0.9 & 0.9 & 0 & 70.9 \\
a15 & 0 & 0.3 & 0.3 & 0 & 52.6 \\
a16 & 0 & 0.4 & 0.4 & 0 & 81.0 \\
a17 & 0 & 0.3 & 0.3 & 0 & 61.5 \\
a18 & 0 & 0.0 & 0.0 & 0 & 64.7 \\
a19 & 0 & 0.8 & 0.8 & 0 & 65.3 \\
a20 & 0 & 0.1 & 0.1 & 0 & 58.0 \\
a21 & 0 & 0.2 & 0.2 & 0 & 74.2 \\
a22 & 0 & 1.2 & 1.2 & 0 & 75.9 \\
a23 & 0 & 0.6 & 0.6 & 0 & 73.6 \\
a24 & 0 & 0.8 & 0.8 & 0 & 77.0 \\
a25 & 0 & 0.4 & 0.4 & 0 & 50.0 \\
a26 & 0 & 0.0 & 0.0 & 0 & 55.3 \\
a27 & 0 & 0.0 & 0.0 & 0 & 40.2 \\
a28 & 0 & 0.5 & 0.5 & 0 & 67.9 \\
a29 & 0 & 0.5 & 0.5 & 0 & 61.4 \\
a30 & 0 & 0.4 & 0.4 & 0 & 65.9 \\
a31 & 0 & 1.0 & 0.0 & 0 & 45.0 \\
a32 & 0 & 2.5 & 0.5 & 0 & 74.6 \\
a33 & 0 & 0.4 & 0.4 & 0 & 59.4 \\
a34 & 0 & 0.4 & 0.4 & 0 & 65.3 \\
a35 & 0 & 1.3 & 1.3 & 0 & 79.9 \\
a36 & 0 & 0.2 & 0.2 & 0 & 63.1 \\
a37 & 0 & 1.1 & 1.1 & 0 & 71.9 \\
a38 & 0 & 0.4 & 0.4 & 0 & 63.5 \\
a39 & 0 & 1.5 & 1.5 & 0 & 66.6 \\
a40 & 0 & 0.0 & 0.0 & 0 & 47.9 \\
\bottomrule
\end{tabular}
\vspace{0.2cm}
    \caption{Results obtained with the 40 problems with the longer turn around windows.  Results are based on 10 runs.}
\vspace{-0.8cm}
    \label{tab:res1}
\end{table}

\begin{table}[h]
\tiny
\centering
\begin{tabular}{l|ccccc}
\toprule
 Instance & Best & Avg. & Missing Staff &Late & \% Evals \\
\midrule
b1 & 0 & 0.0 & 0.0 & 0 & 45.4 \\
b2 & 0 & 0.1 & 0.1 & 0 & 50.3 \\
b3 & 0 & 0.4 & 0.4 & 0 & 69.4 \\
b4 & 0 & 0.2 & 0.2 & 0 & 61.6 \\
b5 & 0 & 0.0 & 0.0 & 0 & 31.2 \\
b6 & 0 & 0.0 & 0.0 & 0 & 40.2 \\
b7 & 0 & 0.0 & 0.0 & 0 & 38.1 \\
b8 & 0 & 0.4 & 0.4 & 0 & 48.4 \\
b9 & 0 & 1.4 & 1.4 & 0 & 66.5 \\
b10 & 0 & 0.6 & 0.6 & 0 & 47.7 \\
b11 & 0 & 0.2 & 0.2 & 0 & 41.9 \\
b12 & 0 & 0.8 & 0.8 & 0 & 45.9 \\
b13 & 0 & 3.6 & 1.6 & 0 & 55.5 \\
b14 & 0 & 0.3 & 0.3 & 0 & 52.8 \\
b15 & 0 & 0.5 & 0.5 & 0 & 64.3 \\
b16 & 0 & 0.0 & 0.0 & 0 & 37.2 \\
b17 & 0 & 0.1 & 0.1 & 0 & 53.8 \\
b18 & 0 & 0.1 & 0.1 & 0 & 47.6 \\
b19 & 0 & 0.4 & 0.4 & 0 & 71.0 \\
b20 & 0 & 0.1 & 0.1 & 0 & 39.6 \\
\bottomrule
\end{tabular}
\vspace{0.2cm}
\caption{Results obtained with the 20 problems that have shorter turn around windows. Results are based on 10 runs}
\vspace{-0.8cm}
\label{tab:res2}
\end{table}

A key result is that, across all 60 problem instances, AERO-EA consistently discovered at least one solution with zero penalties (i.e. no uncovered work orders and all aircraft released within their turnaround windows). This outcome validates the fitness of both the solution representation and the evolutionary operators, and demonstrates the algorithm’s ability to fully resolve even tightly constrained schedules when given sufficient search budget. It should be noted, however, that while the best solution in each case achieved full feasibility, the average fitness values across runs were often greater than zero, especially in more constrained scenarios. This suggests that the search landscape is highly rugged and multi-modal, with a significant proportion of runs converging to local optima, particularly as time windows are compressed.

A comparative analysis between the two batches reveals  patterns. In Batch (a), where maintenance tasks occupied only 35\% of the turnaround window, the problem was notably easier: in 12\% of these relaxed instances, all 10 runs identified a zero-penalty schedule.  Batch (b), representing high-pressure conditions (83\% resource utilisation), saw 25\% of instances where every run was successful. This suggests that the underlying fitness landscape differs between both batches. A consistent observation, across both batches, is that infeasibility, when it occurred, was almost exclusively due to staff allocation conflicts rather than violations of turnaround windows. This highlights the critical role of resource contention, particularly under tight operational margins, and underlines the need for efficient technician allocation and conflict resolution within the EA.

Analysis of the evaluation budget utilisation provides insights into the search dynamics. For almost all instances, a substantial portion of the 200,000 evaluation limit was required to reach an optimal solution, with a lower bound of 35\% and many cases exceeding 70\%. This is unsurprising given the factorial growth of the search space with respect to work packages and staff assignments. The algorithm’s capacity to consistently identify feasible solutions within this relatively modest computational envelope is notable and speaks to the effectiveness of the representation and operators, though the performance could likely be improved by more advanced diversity preservation or hybrid search mechanisms.

The use of multiple independent runs per instance offers further evidence of AERO-EA's robustness. Even in the most challenging scenarios, at least one run per problem achieved the optimal result, underscoring the repeatability and reliability of the evolutionary approach. However, the observed variability in mean fitness between runs also suggests that, for real-world applications where solution quality and consistency are paramount, there may be merit in combining AERO-EA with post-processing local search, ensemble strategies, or adaptive restarts.

From an operational perspective, these results confirm that AERO-EA is capable of reliably generating feasible, high-quality maintenance schedules across a spectrum of real-world-like constraints. In practical airline settings, such capability is critical to minimising aircraft downtime and ensuring regulatory compliance, even when operating near resource or temporal capacity limits.

While this study primarily reports aggregate metrics, future work could greatly benefit from deeper statistical analysis (e.g., distributions of run-time to first feasible solution, spread of fitness values, or sensitivity to parameter variations) and richer visualisation (such as convergence plots, diversity trajectories, or Gantt charts illustrating typical schedules). Such analysis would further elucidate algorithmic behaviour and support practical implementation.

\subsection{Conclusions and Future Work}

This paper has addressed the formidable challenge of aircraft maintenance scheduling, which is characterised by complex temporal dependencies, stringent qualification requirements, and high operational stakes. We have developed and presented AERO-EA, a steady-state evolutionary algorithm that is specifically tailored to generate feasible, high-quality maintenance schedules for fleets of commercial aircraft operating under demanding real-world constraints.

A central contribution of this work lies in the careful problem representation adopted within AERO-EA. The hierarchical chromosome structure allows for flexible reordering of work packages while strictly maintaining the immutability and sequential execution of the constituent work orders. This design ensures that operational realities, such as non-interruptible tasks and mixed qualification requirements, are faithfully captured during both solution construction and evolutionary search. The custom crossover and mutation operators further support the search process by exploring alternative technician allocations and task sequences, all while preserving feasibility with respect to staff certification and availability constraints.

The extensive computational experiments conducted across 60 systematically generated benchmark instances provide strong evidence for the efficacy and robustness of the proposed methodology. Notably, AERO-EA was able to discover a solution with zero penalties for every test instance considered. In other words, for all 60 generated scheduling problems each repeated over 10 independent algorithm runs the system was consistently able to allocate all required maintenance work to suitably qualified staff, within the available turnaround windows, such that every aircraft could be released on time. This result serves as a clear validation that both the representation and the evolutionary operators are well suited for the aircraft maintenance scheduling domain, enabling the system to reliably navigate the vast and highly constrained search space.

Executing the algorithm for 10  runs per instance proved sufficient to reliably locate optimal solutions, as each problem saw at least one run achieve a zero-fitness schedule. This approach not only provides confidence in the repeatability and stability of the algorithm, but also highlights the adequacy of the evaluation budget employed for problems of this scale. While some individual runs particularly under tighter operational constraints incurred penalties due to unallocated work orders, the best run in every case found a feasible and fully staffed schedule. The results also underscore the ability of AERO-EA, to address both relaxed and highly compressed maintenance environments by efficiently coordinating technician resources in the face of significant combinatorial complexity.

These findings have important implications for the deployment of intelligent scheduling tools within the aviation sector. The ability to rapidly generate feasible, high-quality staff allocation plans, even under substantial time pressure, can help airlines optimise both operational efficiency and compliance with regulatory mandates. The demonstrated flexibility of AERO-EA, combined with its capacity for handling heterogeneous staff qualifications, makes it an attractive candidate for integration into digital maintenance planning systems, especially as the sector transitions toward increasingly data-driven and dynamic operations.

Looking forward, several research avenues can further advance the capabilities and applicability of the AERO-EA framework. Foremost among these is the need to evaluate the algorithm’s performance on real-world airline maintenance data, exposing it to genuine operational irregularities, diverse technician rosters, and authentic maintenance task networks, which may reveal limitations or adaptations needed beyond synthetic benchmarks. Incorporating soft constraints related to staff management—such as technician preferences, fatigue limits, or fair workload distribution—will also enhance practical relevance and typically transform the problem into a multi-objective optimisation (MOO) context, requiring trade-offs between operational efficiency and staff welfare. The AERO-EA algorithm itself could be improved through methodological enhancements like adaptive parameter tuning, hybridisation with local search or constraint programming, and advanced evolutionary strategies, all of which may accelerate convergence, increase solution diversity, and reduce computational costs. Additionally, extending the model to address stochastic elements, including last-minute technician absences or emergent maintenance tasks, would foster resilient, real-time adaptive scheduling, while integrating predictive maintenance data from aircraft sensors could facilitate proactive, data-driven planning. Finally, broadening the framework to encompass multi-day planning, varied technician shifts, and collaborative maintenance across airline alliances would significantly expand its real-world impact and utility for aviation maintenance management.
\subsubsection*{Acknowledgements.} The authors gratefully acknowledge the financial support from QOCO Systems Ltd. under project AI-MRO.

\bibliographystyle{splncs04}
\bibliography{bibliography}

\end{document}